# Actionable Interpretation of Machine Learning Models for Sequential Data: Dementia-related Agitation Use Case

Nutta Homdee

Department of Electrical and Computer Engineering, University of Virginia, Charlottesville, Virginia, USA, nh4ar@virginia.edu

Hilda Goins

Department of Industrial & Systems Engineering, North Carolina A&T State University, Greensboro, NC, USA, hgoins@ncat.edu

Azziza Bankole

Department of Psychiatry & Behavioral Medicine, Virginia Tech Carilion School of Medicine, Roanoke, VA, USA, aobankole@carilionclinic.org

Martha S. Anderson

Department of Interprofessionalism, Virginia Tech Carilion School of Medicine, Roanoke, VA, USA, msaconsulting@mail.com

John Lach

Department of Electrical and Computer Engineering, The George Washington University, Washington, D.C., USA, jlach@gwu.edu

## ABSTRACT

Machine learning has shown successes for complex learning problems in which data/parameters can be multidimensional and too complex for a first-principles based analysis. Some applications that utilize machine learning require human interpretability, not just to understand a particular result (classification, detection, etc.) but also for humans to take action based on that result. Black-box machine learning model interpretation has been studied, but recent work has focused on validation and improving model performance. In this work, actionable interpretation of black-box machine learning models is presented. The proposed technique focuses on the extraction of actionable measures to help users make a decision or take an action. Actionable interpretation can be implemented in most traditional black-box machine learning models. It uses the already trained model, used training data, and data processing techniques to extract actionable items from the model outcome and its time-series inputs. An implementation of the actionable interpretation is shown with a use case: dementia-related agitation prediction and the ambient environment. It is shown that actionable items can be extracted, such as the decreasing of in-home light level, which is triggering an agitation episode. This use case of actionable interpretation can help dementia caregivers take action to intervene and prevent agitation.

## KEYWORDS

## 1 Introduction

Machine learning has progressed significantly over the past two decades. Many fields of technology and science, such as finances, speech processing, image classification and health care, have successes in applying machine learning [11]. Machine-learning algorithms have been developed to suit several types of data and learning problems. Generally, these algorithms can be viewed as searching through a large set of possible parameters, guided by training samples, to find a set of parameters that best optimize performance (e.g., highest classification accuracy, least regression error, etc.). Many of the widely-used machine-learning algorithms are considered black-box; Rudin explains that a black-box model could be a function that is too complicated for a human to understand or a proprietary function [19]. Examples of black-box models are the predictive deep learning models that can involve millions of parameters. Deep learning has been implemented in applications such as medicine, finance, and other fields that often require a more complicated model due to the complexity of the data and its learning problems, such as multivariable input data or image/time-series classifications that consider the temporal relationship of the data [23].

However, these implementations of the black-box model can be problematic. The lack of interpretability of the machine learning model can make it difficult for the user to take action or make a decision based on the model result, especially in healthcare applications where decisions can come with severe consequences. Recently, many pieces of research are trying to develop techniques to explain/interpret machine learning algorithms, especially the black-box models. Došilović defines two categories of approaches to model interpretability: (1) integrated interpretation, which traces the logic behind the model, and (2) post-hoc, which extracts information from an already learned model [7]. Samek describes that, currently, machine-learning model interpretation methods often focus on understanding how the model works and why it produces a certain result [20]. This can increase the user's trust in the machine-learning algorithm and can be used to improve the model's performance. However, it does not always provide actionable measures, which are the important component of many applications. For example, if a black box model predicts a certain outcome (the outcome is usually known to the user), what action/decision should the user take to get the desired outcome?

In this work, we focus on the *actionable* interpretation of black-box machine learning models. We also focus on learning problems with sequential/time-series data types that use black box predictive models such as convolutional neural networks or deep neural networks. Our proposed technique uses a combination of information from the model outcome; the already learned model; the used training data; and the newly observed data to extract actionable items, as shown in Figure 1. The actionable items are aimed to help the user make decisions or take actionable measures based on the interpretation; as opposed to many black-box model interpretation studies that focus on getting model information (e.g., model's logic, model parameters, etc.) to improve the performance of the model [20]. The extracted actionable items are related to the time-series behavior of the sequential input data that impact the model result. Examples of these time-series behaviors can be the abnormally low value of the data, the increasing of the data, or sudden changes in the data. The proposed actionable interpretation method is post-hoc, which uses information from the already-learned model without modifying the algorithm of the black-box model. This makes the proposed method robust and can be implemented for most traditional black-box predictive models with sequential/time-series data inputs.

As a case study, we implement our proposed actionable interpretation method with a real-world application of dementia-related agitation and ambient environment causation. Dementia patients express agitated behavior in various ways such as verbal outbursts or aggressive motor behavior [3]. The occurrence of dementia agitation can be affected by the ambient environment, such as the patient's long exposure to high ambient sound level [10] or bright light [12]. We use a black box prediction model to predict upcoming dementia agitations based on the in-home ambient environment around the dementia patient. Then, we implement our actionable interpretation method to determine the potential ambient environment that causes the patient's agitation and extract actionable items related to the ambient environment's behavior. An example of an actionable item, in this use case, could be to notify the in-home dementia caregiver to turn the lights on because the ambient light level is decreasing (e.g. the sun is setting), which has triggered agitation episodes in the past.

The main contributions of this work are:

1. Actionable model interpretation technique that uses the prediction result, the already learned model, the used training data, and the newly observed data without modifying the black-box model.
2. Sequential/time-series analysis techniques, using permutation importance and cross-correlation, for the model's predictors' importance and time-series behavior extraction.
3. Per-observation interpretation does not require the whole testing dataset to interpret models, which is suitable for applications that require real-time action/decision.
4. Actionable interpretation in a use case: ambient environmental causation for dementia-related agitation.

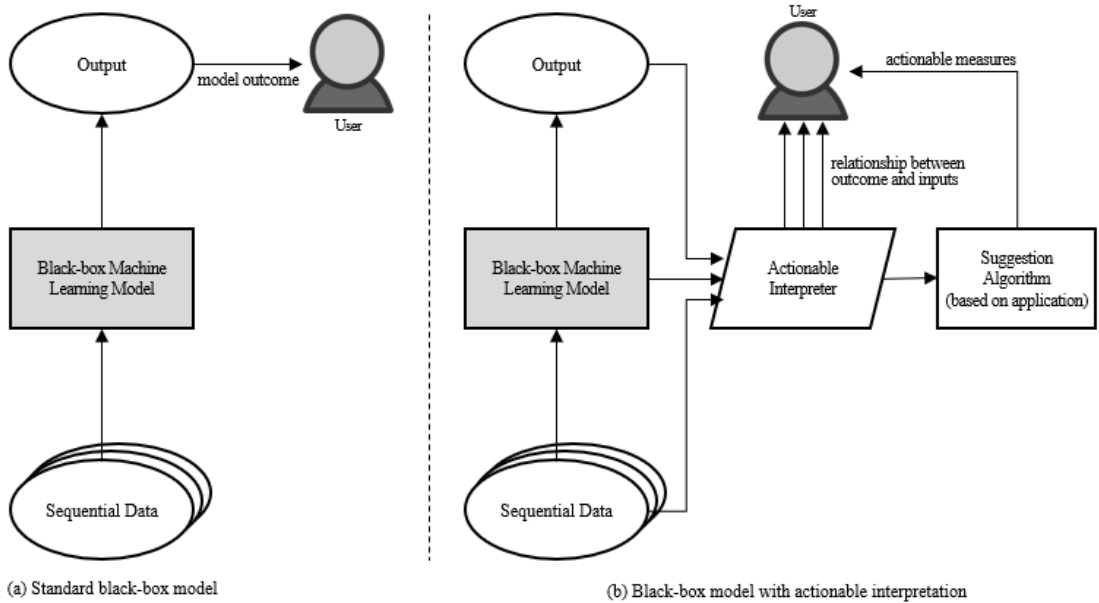

Figure 1: Standard black-box machine learning model vs. model with actionable interpretation. (a) Standard machine learning model: users only receive a model outcome, difficult to make a decision or take action based on the outcome only. (b) A model with the proposed actionable interpretation: users receive suggested actions based on the already learned model, the used training dataset, and the model outcome. No modification is made to the black-box model.

## 2 Related Work

Recently, many works have been conducted in the model interpretation area. In this section, we discuss the related work regarding machine learning interpretability and our approach, actionable interpretation. The proposed approach focuses on interpreting black-box machine learning models to extract actionable items (e.g., the ambient environment's correlation to the prediction of an upcoming dementia-related agitation event). This actionable interpretation technique is aimed to help the user make decisions or take actions from machine learning models rather than interpreting models to validate outcomes or improve model performance.

Approaches to machine learning interpretation focus on explaining the model's logic by tracing through parts of the model [16]. Došilović categorized this type of model interpretation as an integrated interpretation in which the interpretability comes from the transparency of the model [7]. This interpretation approach can be limited to simpler models such as linear models and decision trees. In the decision tree models, the logic in the model such as tree split gains and sample weights can be used to explain the model [22]. Other works make sense of complex models by modifying the model to extract interpretable information. Martens extract rules from support vector machines (SVM) to make the model interpretable for credit scoring [17]. Choi modified the recurrent neural network (RNN) to generate coefficients at certain stages of the network [2]. The generated coefficient can be used to calculate input contribution to the model's output. These works often rely on models with low complexity or require certain modifications to learning models that make it specific to a certain situation. Also, the integrated interpretations often give insights that can be used for the model's verification and improvement; but not necessarily appropriate in the extraction of actionable items.

Another category of machine learning model interpretation is post-hoc interpretation, which extracts the model's information from the already learned model [7]. Krause validates predictors to the diabetes risk prediction by implementing partial dependence plots which show the changes in the model's output when the predictor's value changes [14]. Colubri evaluated the impact of the predictors to the prediction outcome of Ebola patients by analyzing the already trained model to rank the predictors [4]. In both of the mentioned studies, the predictor interpretation is done using all data samples – training samples, testing samples - to see which predictors contribute a high impact on the prediction outcome. This method may not be suitable to be implemented in a real-time situation, especially in applications that required real-time action/decision. In this work, we also address the real-time implementation by a predictor interpretation technique per-observation. Still, an advantage of these post-hoc model interpretation approaches in the robustness because they can be implemented to most traditional machine-learning models. The proposed actionable interpretation uses the post-hoc predictor interpretation to find impactful predictors that can be used to extract actionable items from the models.

Many complex, black-box, machine learning models have been implemented for sequential data such as video records and time-series data. In this work, we focus on the actionable interpretation of such models. The interpretation of models with sequential data can be complicated, as traditional model interpretation may lack insights on the temporal relationship in the sequential data. Norgeot conducted ambulatory outcome forecasting from time-series electronic health record features [18]. In the study, the features are extracted from window sequences in the time-series data. Then, model interpretation is done using the permutation importance scoring technique to rank the important features. This interpretation technique still only provides the user with information regarding the importance of the predictors. It does not give insight into the temporal behavior of the time-series predictors. Information on temporal behavior (e.g., increasing, decreasing) of time-series inputs could be directly used to take actionable measures such as disrupting unwanted time-series behavior. In this work, as part of the actionable model interpretation, we also provide techniques to analyze the behavior of the time-series/sequential predictors, which is crucial in the extraction of actionable measures for the models' users.

## 3 Methods

In this section, the proposed method, actionable interpretation for the black-box predictive model, is explained. Our actionable interpretation method focuses on the extraction of items that the model's users can take action upon. The extracted actionable items are the behavior of the model's input data (sequential/time-series data). As an example, in our use case, dementia-related agitation prediction from the ambient environment, the extracted actionable items are the ambient environmental conditions and dynamics that are likely to trigger an agitation event. Such behaviors could be low-light-level, sudden changes in ambient sound level, or the decreasing in-home temperature/humidity. By learning the potential trigger of an upcoming agitation, the user (e.g. in-home dementia caregiver) can intervene or de-escalate agitation episodes.

An overview architecture of the actionable interpretation methods is shown in Figure 2. The methods consist of: (1) rate-of-change (ROC) calculation, (2) predictors ranking for sequential/time-series predictors, and (3) actionable items extraction. The ROC of a time-series input is utilized to help the actionable interpretation identify appropriate time-series behavior (e.g. stationery, changing). The ROC calculation is implemented both during the model training stage and when using the model for predictions. When there is a new observation data, the already trained model uses both the observation data and its ROC as inputs. The outcome of the prediction model – model prediction (e.g. classification output) and prediction probability – is used in the predictor ranking. The predictor ranking is used to find impactful predictors; at this stage, the predictor ranking also provides information on whether the predictor's value of its ROC has more impact on the prediction outcome. This information is used for actionable items extraction which extracts behaviors of the impactful predictors. All of the mentioned techniques can be used to extract actionable items without any modification to the model, making them implementable to most black-box machine learning algorithms.

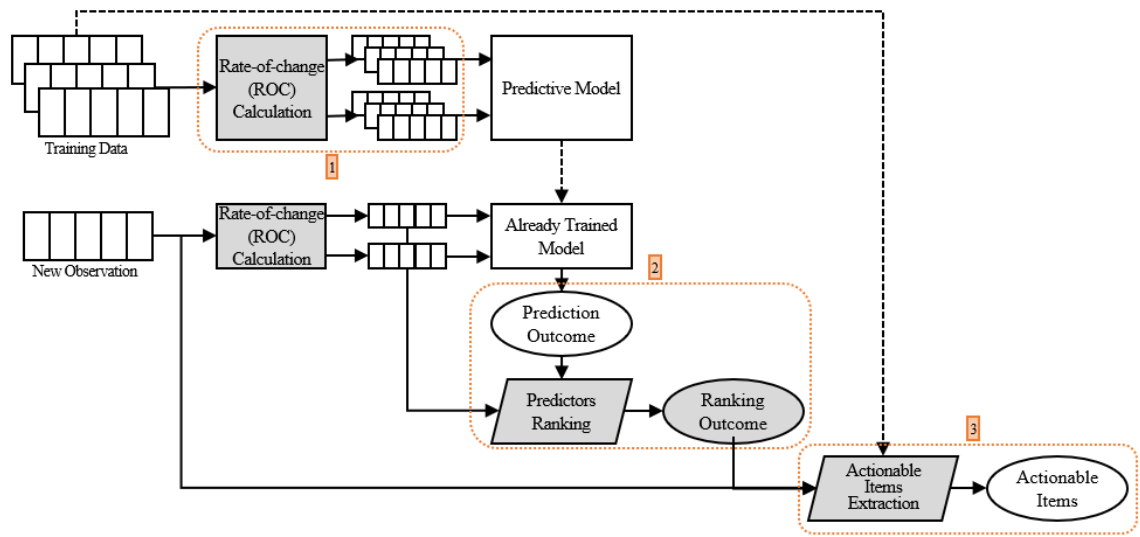

**Figure 2: Actionable interpretation overview architecture: The interpretation uses an already trained model, used training data, and model's outcome to extract actionable items and provide an actionable measure to users. Components of the actionable interpretation for black-box models are (1) rate-of-change calculation, (2) predictors ranking, and (3) extraction of actionable items.**

## 3.1 Calculating rate-of-change (ROC)

The goal for the proposed actionable interpretation is to find the time-series behavior of the impactful predictors. The predictor's behavior can be statistically stationary or trend-differencing [24]. Thus, the calculation of the rate-of-change (ROC) in the time-series data is used to better understand whether it is the stationary behavior of the data (e.g. value too low) or the changes in the data (e.g. increasing, decreasing) that impact the prediction outcome the most. Before the model is trained, the ROC of the input/training data needs to be calculated. Then, both the training data and their ROC are used as inputs to train the model, as shown in Figure 2. When the already trained model is used, e.g. for prediction, the calculation of ROC is also required.

In this work, we use the absolute of a gradient of time-series data to calculate the ROC. Let a time-series array $X = X_0, X_1, X_2, \ldots, X_i, \ldots, X_{n-1}, X_n$. The gradient of a time-series array $X$ is defined as:

$$\nabla X_i = \begin{cases} X_1 - X_0; & i = 0 \\ (X_{i+1} - X_{i-1})/2; & 1 \leq i \leq n-1 \\ X_n - X_{n-1}; & i = n \end{cases} \quad (1)$$

Thus, the ROC of a time-series data $X$ is defined as shown in the equation below:

$$\text{ROC}_X = |\nabla X| \quad (2)$$

In equation (1), time-series data with high noise can result in a high fluctuation in the ROC. Therefore, a noise reduction filter is recommended before calculation of the ROC. Note that the requirement of a noise reduction filter depends on the application and its time-series data. In our use case, dementia-related agitation prediction from the ambient environment, a Gaussian filter is used to reduce the noises which come from the sensors used to collect ambient environmental data. More information is explained in the use case section. A graphical example of a time-series data and its ROC is presented in Figure 3.

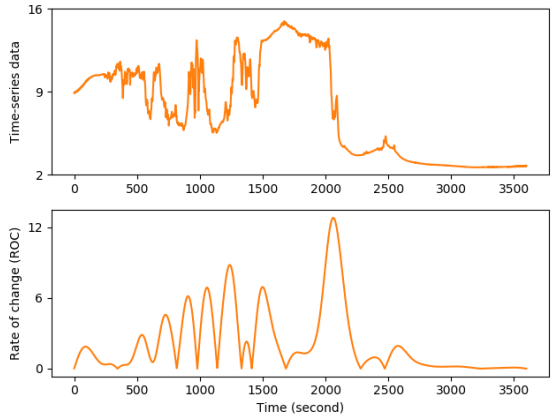

**Figure 3. Time series plot of (a) original time-series data and (b) rate-of-changes**

## 3.2 Predictor ranking

Predictor ranking is used to determine impactful predictors or predictors which highly contribute to the outcome/decision/predictions of the model. From the outcome of the prediction model, we record the prediction probability. The prediction probability can be acquired using methods depending on the machine-learning algorithm such as the output layer's value before the classification activation function (Neural Network-based algorithm), or the value of the last node before leaf nodes (Tree-based algorithm).

After the prediction probability is acquired, the predictor ranking uses the idea of permutation importance on time-series data [1]. The permutation importance is a post-hoc interpretation method in which the calculation is done after a model is trained, and does not require any changes in the model. In this work, we use permutation importance to determines the predictors that have a high impact on predictions by disarranging one predictor at a time (note that predictors can be a normal time-series data or its ROC); The disarranged predictor will lose any time-series pattern which contributes to the prediction of the model. Then, the already trained model is used for prediction again, but with a dataset with one disarranged predictor. A new prediction probability, from a data with a disarranged predictor, is recorded and compared with the original prediction probability. This process of disarranging one predictor at a time, recording new prediction probability, and compare with the original probability is repeated until every permutation of one disarranged predictor is achieved. The predictor which, when disarranged, has a high impact on the prediction probability (compared to the original) is determined to be the impactful predictors. This predictor ranking technique uses the prediction probability of an observation to rank predictor importance, as opposed to the traditional approach which uses a set of observations [1]; Making this ranking method implementable for real-time application where the model only sees one observation at a time.

## 3.3 Actionable items

Actionable items are extracted from predictors which are determined to have a high impact on the model's predictions. In this work, we learn the time-series behavior of the high impact predictors and extract actionable items from the behavior. For example, the proposed actionable interpretation can learn the model's prediction is impacted by the decreasing of a predictor. Then, based on the current application, we can produce an appropriate actionable measure to prevent/stop the decreasing of the predictor. In this work, the detection of time-series behaviors can be divided into two categories: (1) statistically stationary, and (2) differencing or the changing in the time-series data [24].

During predictor ranking step, if the predictor with high impact to the predictions is the original time-series (not its ROC), stationary behavior will be extracted. First, the action interpretation learns basic statistics from the used training data which are the means and standard deviations of each predictor. The learned statistics from the used training data are compared with the average value of the high impact predictor. The comparison provides information on the predictor's stationary behaviors. Possible stationary time-series behaviors are the abnormally low value and abnormally high value. If the average value of the high impact predictor is significantly low compared to the mean and standard deviation of the same predictor, from the used training data, the behavior is inferred as abnormally low value.

On the contrary, if the predictor ranking informs that the ROC of a predictor has a high impact, the actionable items will be extracted from the time-series differencing behavior instead. In this work, we focus on time-series differencing behaviors which are: increasing, decreasing, or sudden changes in the data. To learn the differencing behaviors, a cross-correlation between the original time-series predictor and our cross-correlation filters is computed. Let a time-series array $X = X_0, X_1, \ldots, X_i, \ldots, X_{n-1}, X_n$ and a cross-correlation filter $g = g_0, g_1, \ldots, g_i, \ldots, g_{n-1}, g_n$ For a better cross-correlation comparison, we normalized $X$ with its mean $\bar{X}$ and standard deviation $\sigma_x$ [21]. The cross-correlation filters are designed to have the same size as the time-series data and also normalized. The normalized cross-correlation between $X$ and $g$ is defined as:

$$(X * g)_i = \sum_{m=-\infty}^{\infty} \frac{(X_{m-i} - \bar{X})}{\sigma_x} \times \frac{(g_m - \bar{g})}{\sigma_g} \qquad (3)$$

Cross-correlation represents the similarity between the predictor's pattern and the filters which are: increasing filter, decreasing filter, and sudden changes filters (positive and negative sudden changes). The cross-correlation filters are shown in Figure 4. After the cross-correlation is computed, we extract the maximum value. The point of maximum value represents when the two data array (time-series data and the filter), are most similar. The differencing time-series behavior with the highest cross-correlation maximum value is identified to be the impactful behavior.

The extraction of actionable items can provide insights on the action/decision that the user can take which will impact the predictions. The exact course of action is usually depending on the application which is discussed in the use case application section.

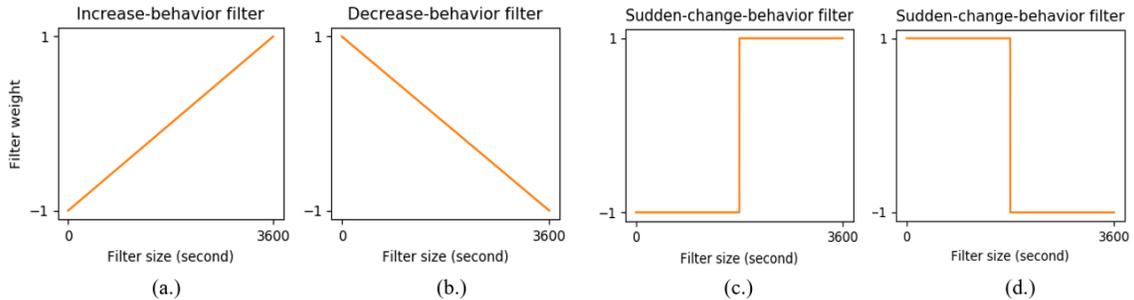

(a.) Increase-behavior filter (b.) Decrease-behavior filter (c.) Sudden-change-behavior filter (d.) Sudden-change-behavior filter

**Figure 4. Cross-correlation filters: (a) increasing behavior filter, (b) decreasing behavior filter, and (c, d) sudden changes behavior filters.**

# 4 Use Case: Dementia-related Agitation and Ambient Environment

We use the proposed actionable interpretation to extract actionable items ambient environment and dementia-related agitation prediction. Persons with dementia (PWD) suffer from cognitive impairments which often lead to difficult daily lives. The difficulty related to the impairments can cause the PWD to express agitated behavior such as verbal outbursts, or aggressive motor behaviors which can be called dementia-related agitation. Kales states that the occurrence of agitation can be unpredictable or be affected by the ambient environment around PWD [12]. Many studies have shown the causation between the ambient environment and PWD. Joosse from the College of Nursing, University of Wisconsin-Milwaukee, suggests that dementia agitation has a significant correlation with the level of sound in the environment (in this case, nursing homes) [10]. Another study by Van Hoof observed that high-intensity bright light can affect restlessness behavior and increase the frequency of agitation occurrences of institutionalized older adults with dementia [9].

## 4.1 Real-world dataset

We collected ambient environmental data in the homes of PWD using our developed integrative sensing system; through a collaboration with dementia experts from the Virginia Tech Carilion School of Medicine [8]. The system consists of room-level sensing nodes mounted with environmental sensors that collect light level, in-home temperature, humidity, air pressure, and ambient noise level. All environmental data is sequentially sampled once per second (1 Hz) except for the ambient noise level which is sampled at 10 Hz then averaged down to 8 Hz to remove any distinguishable conversation. The system has been deployed to collect data in real dementia patient and caregiver (CG) homes. Each takes two-months during which the system passively collects the environmental data. To gather information about the occurrences of the in-home dementia-related agitation, CGs use a tablet survey application to label each agitation episode. On average, we received 6 agitation labels per week or around 48 agitation labels for each two-month deployment. Figure 5 shows an example period of the collected ambient environmental data and the CG provided agitation label.

To learn ambient environmental data patterns that correlate to the occurrences of dementia-related agitation, we used the data immediately prior to each agitation label (CG provided). One-hour segments of ambient environmental data are extracted to be used as predictors. The one-hour segment ranges from 72-minutes before each agitation label to 12-minutes before. Ambient environmental data around the time of agitation label has not been used, because we observed that there are certain changes in the data during most agitation episodes. For example, we often see high-changes of the ambient noise level during the time of agitation; which can be explained as verbally agitated behavior or verbal activities between PWD and CG. Since we are interested in the effect of the ambient environment on PWD which leads to agitation, we excluded the data at the time of agitation. The time-segment period of the ambient environment that the predictive model used to train patterns is shown in Figure 5. Additionally, the time-of-day is also added as one of the predictors. The time-of-day can be a useful agitation predictor. Khachiyants explain that many PWDs often show agitated behaviors at a similar time of the day which is called sundowning syndrome [13].

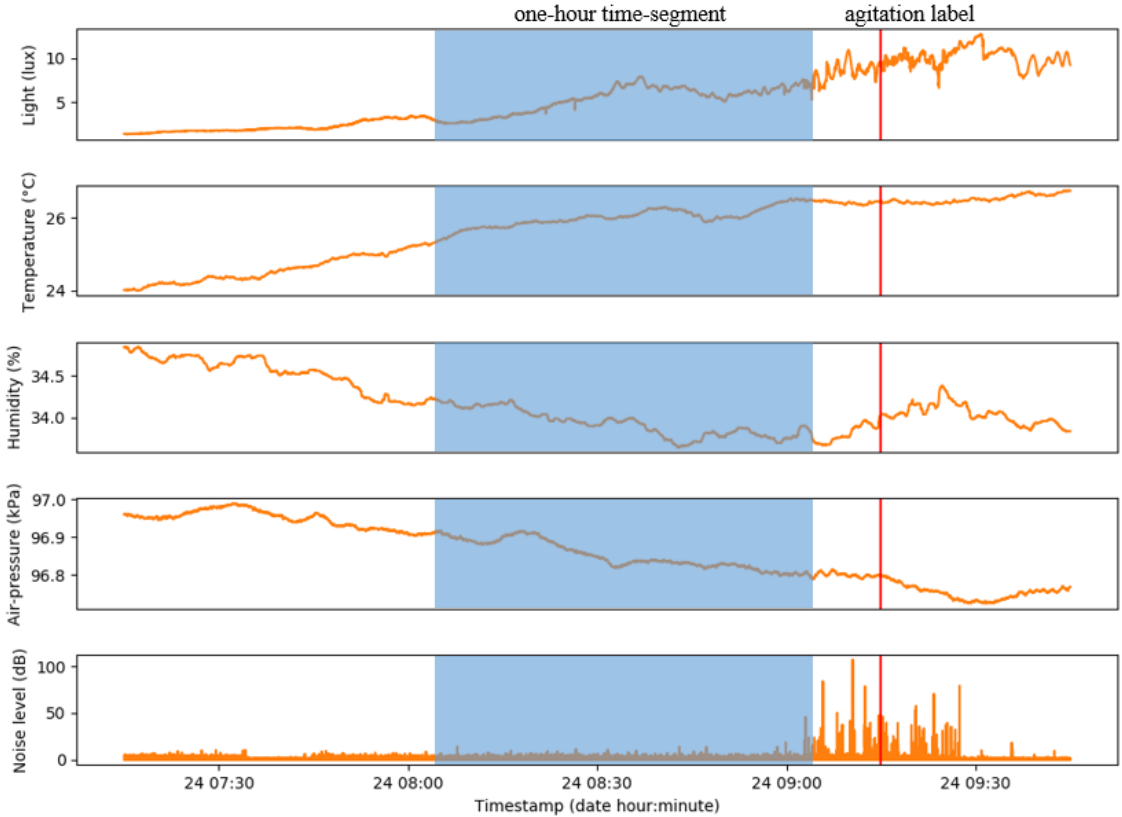

**Figure 5.** A period of ambient environmental data collected by the sensor node at PWD home. The ambient environmental data, from top to bottom, are (from top to bottom) light level (lux), ambient temperature (°C), indoor humidity level (% relative humidity), air-pressure (kPa), and ambient noise level (dB). The red vertical line shows a dementia-related agitation event label provided by the PWD's CG. The highlighted area represents a time-segment period of the ambient environment that our predictive model used to train patterns that lead to an agitation event.

## 4.2 Predictive modeling framework

This learning problem required a model for multi-variable pattern recognition (multiple ambient environment time-series) and a model considering temporal relationship. Thus, we use a Convolutional Neural Network (CNN) to predict upcoming agitation episodes based on the one-hour time segments of the ambient environmental time-series. CNN has six input heads, five for each of the ambient environment, and one for time-of-day. For the dataset, we use the CG provided agitation labels as positive labels. Time segments of the negative labels (non-agitation) are randomly picked using the time when no agitation has been reported. The ratio between agitation observations and the randomly chosen non-agitation periods is 1:4 due to the highly imbalanced nature of the real-world dementia-related agitation. The training set consists of 70% of all the observations.

Using complicated prediction models, such as CNN, with an imbalanced and small dataset, problems such as model overfitting and one-class prediction are likely to occur. To counter these problems, we combine the CNN with a regularization algorithm which is well-performed on real-world problems. We implemented a gradient boosting algorithm for the regularization. The gradient boosting has parameters to handle a small/imbalanced dataset [1]. The idea is to learn patterns in the ambient environment using CNN, then utilize the gradient boosting which is more suitable for a small and unbalanced dataset. We achieve this by: first, train/optimize the CNN to learn patterns in the ambient environment; then, extract the Fully Connected Layer (FCL) of the trained CNN; use the extracted FCL of CNN as inputs to the gradient boosting algorithm to handle small/imbalanced dataset. The optimized parameters of the CNN are: 16 CNN filters for each input head, filter's sizes are optimized differently for each ambient environment, 30-seconds CNN pooling size – an ambient environmental time-series has 3600-seconds in size, and 50% CNN drop rate. For the gradient boosting, five-fold cross-validation has been implemented and the optimized parameters are: 12 max tree depth, minimum child weight is set to 1, regularized parameter to 0.005 – for a small dataset, and positive/negative label weight is set to 1:4 ratio.

After the CNN/gradient boosting model is trained, we implement the proposed actionable interpretation technique using the rest of the observations. For each observation that the model classifies as positive - the model predicts an upcoming agitation – we acquire the prediction probability and extract actionable items from the predictors. Each predictor (ambient environmental time-series and their ROCs) is disarranged, used to predict agitation again, and compared with the original prediction probability for predictor ranking. After that, the time-series behavior of the important predictor is extracted for the actionable item. For example, an ambient environmental observation is predicted to lead to agitation; the ROC of light level is the impactful predictor; the decreasing of the light level is the extracted behavior. In this case, we can conduct an actionable item to turn the light on to stop the decreasing light level.

## 5 Results

The proposed actionable interpretation method is used with the dementia-related agitation and ambient environmental use case. The actionable interpretation includes the calculation of ROC, predictor ranking techniques for time-series predictors, and actionable items extraction. By implementing the proposed technique, we extract ambient environment behaviors that correlate to the occurrences of dementia-related agitation in the homes of PWD.

Figure 6 shows the implementation of the actionable interpretation of one of the observations. The ambient environmental data, as shown in Figure 6a, is predicted to cause an upcoming agitation by the predictive model. The predictor ranking technique determines important predictors by disarranging one predictor at a time and comparing the prediction probabilities. The predictor importance scores (shown in Figure 6b) are computed by subtracting the prediction probability of a disarranged predictor to the original prediction probability. If a predictor has a high impact on the agitation prediction, when that predictor is disarranged, the prediction probability will alter drastically. This results in a high predictor importance score. Figure 6b shows that, in this observation case, the ROC of light has the highest impact on the prediction of dementia-related agitation. Since the ROC of light is the important predictor, the actionable item is extracted from the differencing time-series behavior of the light data. Targeted differencing time-series behaviors are increasing, decreasing, and sudden changes in the data. Figure 6c shows the result of the most likely behavior which is computed by cross-correlating the light data with cross-correlation filters. The similarity scores are then normalized mean value of the cross-correlation result. The result shows that the sudden changes in the light data are the most likely cause of the upcoming agitation episode. Thus, the actionable measure could be to close the window (if during daytime) or stop activities which cause fluctuations of light (e.g. CG walks in-out of bedroom).

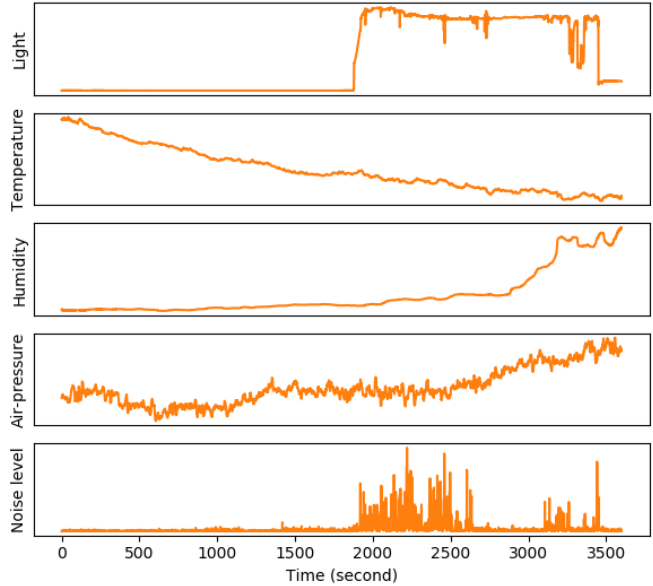

(a.)

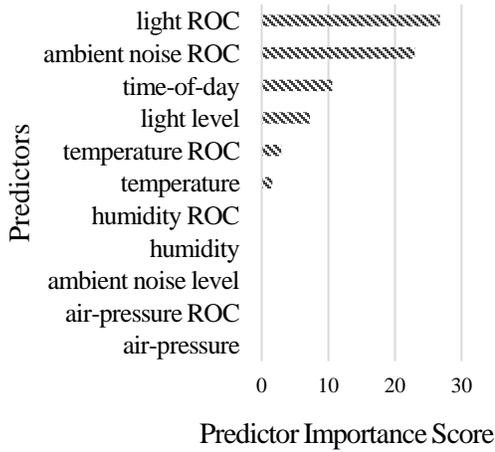

(b.)

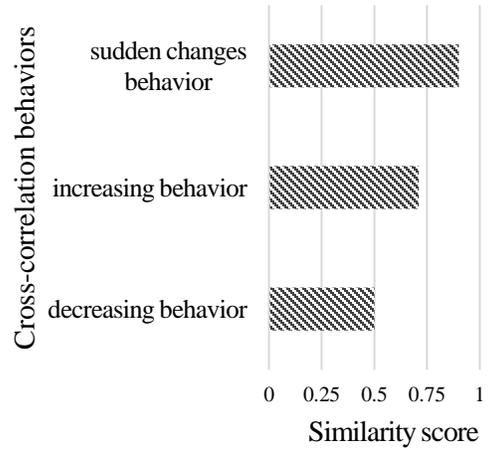

(c.)

**Figure 6. Actionable interpretation of dementia-related agitation on one observation. (a.) An ambient environment observation (one-hour period) that results in a positive prediction classification (upcoming agitation is predicted). (b.) Predictors ranking result showing rate-of-change (ROC) in the light as the most impactful predictor. Predictor importance score is the subtraction of disarranged prediction probability from the original probability. (c.) Cross-correlation results which present that the light ROC is showing sudden changes behavior.**

The summary of the implementation of the dementia-related agitation use case is also presented in Table 1. Here, the actionable interpretation technique has been used to learn ambient environmental behaviors that can potentially cause upcoming dementia agitation. In the table, five two-month deployments of the real-world data collection of in-home ambient environment and dementia agitation labels are shown. An average of 48 agitation labels were reported by the CG. This makes an averaged total observation (agitation labels and non-agitation labels) of 240 observations per deployment. The agitation prediction performance of the standalone CNN model and the CNN/regularization model is also presented with accuracy, weighted F1 score, and standard deviation. The weighted F1 score is chosen as a performance metric to account for the imbalance in the ratio between the number of agitation observations and the number of non-agitation observations. The standard deviation, computed from the different accuracies and weighted F1 scores at the five-fold cross-validation, represents the overfitting of the model; a higher standard deviation demonstrates that the model is only trained to a very specific pattern in the data and not robust enough for most observations. The performance result shows that, on average across deployments, the CNN/regularization model has 14% better accuracy and a 22% better weighted F1 score. Also, the standard deviation performance of the CNN/regularization is significantly lower compared to the CNN model. This shows that the CNN/regularization does not overfit even with the small and unbalanced dataset. Table 1 also shows the common ambient environment and its time-series behaviors that the actionable interpretation determined to be impactful to the agitation predictions. The percentage of the common ambient environmental cause for agitation shows the number of times the predictor (e.g., light level, audio ROC) is determined to be the most impactful predictor out of all observations in which the agitation is predicted. The last column shows common ambient environment behavior that the actionable interpretation detects as potential causes for upcoming agitations, which then become the basis for recommended actions that could be taken to prevent an agitation episode.

**Table 1: Dementia-related agitation prediction performance and actionable interpretation.**

| Deployment # | Observation size (#agitation/non-agitation) | Algorithm Performance (Percentage ± *SD.) | | Actionable Interpretation | |
|---|---|---|---|---|---|
| | | CNN | CNN+ regularization | Common ambient environmental cause for agitation | Common actionable items |
| 1 | 81/300 | Accuracy: 77±7.02 Weighted F1:69±5.26 | Accuracy:91±2.63 Weighted F1:90±2.17 | Time-of-day: 42% Noise level: 33% | High noise level |
| 2 | 16/65 | Accuracy: 75±11.18 Weighted F1:63±8.14 | Accuracy:84±4.2 Weighted F1:80±3.1 | Audio ROC: 60% | Audio sudden changes |
| 3 | 48/192 | Accuracy: 71±8.17 Weighted F1: 65±6.68 | Accuracy: 93±2.52 Weighted F1: 91±2.11 | Light ROC: 52% Noise ROC: 39% | Light level increasing Noise sudden changes |
| 4 | 24/96 | Accuracy: 80±11.02 Weighted F1: 61±7.00 | Accuracy: 89±3.44 Weighted F1: 83±2.91 | Light ROC: 47% Temp** ROC: 44% | Light level decreasing Temperature decreasing |
| 5 | 73/292 | Accuracy: 75±6.44 Weighted F1: 67±5.16 | Accuracy: 90±2.49 Weighted F1: 89±2.02 | Light level: 65% Time-of-day: 29% | High light level |

*SD: Standard Deviation, **Temp: temperature

# 6 Conclusions and Discussion

In this work, an actionable interpretation of black-box machine learning models for sequential/time-series predictors is presented. This differs from interpretability methods used in other works that help to validate and optimize models. The proposed technique aims to assist the model's user in taking actionable steps or to make a decision by extracting actionable items. The actionable items are based on the time-series behavior of the predictors that have a high contribution to the predictions. The actionable interpretation is robust and can be implemented in most traditional black-box machine learning models. The proposed technique achieves this because it is a post-hoc interpretation that utilizes an already learned model, used training data, and post-trained computations without any modification to the model itself. Sequential/time-series data analysis techniques are also utilized to acquire actionable items from a black-box model. This work shows that this technique can interpret the predictors of each observation. This makes the proposed technique appropriate in real-time applications.

The actionable interpretation technique is implemented on a use case of dementia-related agitation prediction from ambient environmental time-series data. Here, the proposed method can enable the user (i.e., in-home dementia CG) to be notified of an upcoming PWD agitation episode and also actionable measures to potentially prevent the episode from occurring. Table 2 shows examples of the actionable measures that are generated based on the extracted actionable items and are developed by our team's dementia experts (Anderson and Bankole). The intervention/prevention of in-home PWD agitations is essential, as dementia-related agitation can cause high burden and stress to the CG; and the CG burden associated with dementia agitation has been reported to be one of the principal factors prompting the institutionalization of community-dwelling PWD [6, 15].

**Table 2: Examples of actionable measures for dementia-related agitation intervention**

| Ambient environmental behaviors* | Agitation intervention suggestions for CG |
|---|---|
| High ambient noise level | Turn off the television/radio before you start talking to minimize background noise and distraction. |
| Sudden changes in ambient noise level | Consider lowering sound in the room to decrease agitation. |
| Low ambient noise level (no stimulation) | Provide appropriate levels of stimulation e.g. talking in a conversational tone, calming music. |
| Decreasing light level | To prevent sundowning, increase lights at least one hour before typical onset. |
| High light level | When hyperactivity is an issue, bright fluorescent lights should be turned down. |
| Sudden changes in ambient light level | Consider lowering or increasing light in the room to decrease agitation. |
| Decreasing in-home temperature | Make sure basic needs are met: temperature, comfort. Close windows and turn-on AC if needed. |

*Time-series behavior that the actionable interpretation determines as high impact to the agitation prediction

The proposed actionable interpretation also contributes to the studies of causality in machine learning. Usually, a machine learning model with good accuracy represents a high correlation between the input(s) and the outcome/target. It does not validate the causality between the inputs and the model outcome. The nature of proving causality in machine learning algorithms is still difficult in a lot of cases, and common sense or domain experts' opinion is used to infer the causality. Accordingly, Das states that "The general agreement in the statistics community is that you cannot prove a causal effect at least without performing an experiment" [5]. The actionable interpretation can be a tool that provides users with actionable measures. Then, actionable measures can be directly used in experiments to prove the causality effect of the input(s) and the outcome.

## ACKNOWLEDGMENTS

This project is supported in part by the National Science Foundation under Grant IIS-1418622.